\documentclass{article} % For LaTeX2e
\usepackage{iclr2016_conference,times}
\usepackage{hyperref}
\usepackage{url}
\usepackage{natbib}
\usepackage{amsmath}
\usepackage{graphicx}
\usepackage{subfig}
\usepackage{pbox}
\usepackage{placeins}

\title{Regularizing RNNs by Stabilizing Activations}

\author{David Krueger \& Roland Memisevic \\
Department of Computer Science and Operations Research\\
University of Montreal\\
Montreal, QC H3T 1J4, Canada \\
\texttt{\{david.krueger@umontreal.ca, memisevr@iro.umontreal.ca\}} \\
}

\newcommand{\norm}[1]{\left\lVert#1\right\rVert}

\begin{document}

\maketitle

\begin{abstract}
	We stabilize the activations of Recurrent Neural Networks (RNNs) by penalizing the squared distance between successive hidden states' norms.  
	This penalty term is an effective regularizer for RNNs including LSTMs and IRNNs, improving performance on character-level language modeling and phoneme recognition, and outperforming weight noise and dropout.
    We achieve competitive performance (18.6\% PER) on the TIMIT phoneme recognition task for RNNs evaluated without beam search or an RNN transducer.
	With this penalty term, IRNN can achieve similar performance to LSTM on language modeling, although adding the penalty term to the LSTM results in superior performance.
	Our penalty term also prevents the exponential growth of IRNN's activations outside of their training horizon, allowing them to generalize to much longer sequences.
\end{abstract}

\section{Introduction}
Overfitting in machine learning is addressed by restricting the space of hypotheses (~i.e.~ functions) considered.
~This can be accomplished by reducing the number of parameters or using a regularizer with an inductive bias for simpler models, such as early stopping.
More effective regularization can be achieved by incorporating more sophisticated prior knowledge.
Keeping an RNN's hidden activations on a reasonable path can be difficult, especially across long time-sequences.
With this in mind, we devise a regularizer for the state representation learned by temporal models, such as RNNs, that aims to encourage stability of the path taken through representation space.
Specifically, we propose the following additional cost term for Recurrent Neural Networks (RNNs):

$$
\beta \frac{1}{T}\sum_{t=1}^T (\norm{h_t}_2 - \norm{h_{t-1}}_2)^2
$$

Where $h_t$ is the vector of hidden activations at time-step $t$, and $\beta$ is a hyperparameter controlling the amounts of regularization. 
We call this penalty the \emph{norm-stabilizer}, as it successfully encourages the norms of the hiddens to be stable (i.e.~ approximately constant across time).
Unlike the ``temporal coherence'' penalty of \citet{RobotPriors}, our penalty does \emph{not} encourage the state representation to remain constant, \emph{only its norm}.

In the absence of inputs and nonlinearities, a constant norm would imply orthogonality of the hidden-to-hidden transition matrix for simple RNNs (SRNNs). 
However, in the case of an orthogonal transition matrix, inputs and nonlinearities can still change the norm of the hidden state, resulting in instability.  
This makes targeting the hidden activations directly a more attractive option for achieving norm stability.
Stability becomes especially important when we seek to generalize to longer sequences at test time than those seen during training (the ``training horizon'').

The hidden state in LSTM \citep{LSTM} is usually the product of two squashing nonlinearities, and hence bounded.  
The norm of the memory cell, however, can grow linearly when the input, input modulation, and forget gates are all saturated at 1.
Nonetheless, we find that the memory cells exhibit norm stability far past the training horizon, and suggest that this may be part of what makes LSTM so successful.

The activation norms of simple RNNs (SRNNs) with saturating nonlinearities are bounded.
With ReLU nonlinearities, however, activations can explode instead of saturating.
When the transition matrix, $W_{hh}$ has any eigenvalues $\lambda$ with absolute value greater than 1, the part of the hidden state that is aligned with the corresponding eigenvector will grow exponentially to the extent that the ReLU or inputs fails to cancel out this growth.

Simple RNNs with ReLU \citep{IRNN} or clipped ReLU \citep{Baidu} nonlinearities have performed competitively on several tasks, suggesting they can learn to be stable.  
We show, however, that IRNNs performance can rapidly degrade outside of their training horizon, while the norm-stabilizer prevents activations from exploding outside of the training horizon allowing IRNNs to generalize to much longer sequences.
Additionally, we show that this penalty results in improved validation performance for IRNNs.
Somewhat surprisingly, it also improves performance for LSTMs, but not tanh-RNNs.

%%%%%%%%%%%%%%%
% PREVIOUS WORK
To the best of our knowledge, our proposal is entirely novel. 
\citet{Razvan} proposed vanishing gradient regularization, which encourages the hidden transition to preserve norm in the direction of the cost derivative.
Like the norm-stabilizer, their cost depends on the path taken through representation space, but the norm stabilzer does not prioritize cost-relevant directions, and accounts for the effects of inputs as well.
A hard constraint (clipping) on the activations of LSTM memory cells was previously proposed by \cite{sak2015}.
\citet{Baidu} use a clipped ReLU, which also has the effect of limiting activations.  
Both of these techniques operate element-wise however, whereas we target the activations' norms.
Several other works have used penalties on the difference of hidden states rather than their norms \citep{RobotPriors, semantic}.
Other regularizers for RNNs that do not target norm stability include weight noise \citep{WN} and dropout \citep{pham13, dropout1, zaremba14}.

%%%%%%%%%%%%%%%%%%%%%%%%%%%%%%%%%%%%%%%%%%%%%%%%%%%%%%%%%%
\section{Experiments}

\subsection{Character-Level Language modeling on PennTreebank} \label{ptb}

We show that the norm-stabilizer improves performance for character-level language modeling on PennTreebank \citep{ptb} for LSTM and IRNNs
\footnote{As in \citet{IRNN}, we initialize $W_{hh}$ to be an identity matrix in our experiments}, but \emph{not} tanh-RNNs.
We present results for $\beta \in \{0, 50, 500\}$. 
We found that values of $\beta > 500$ could slightly improve performance, but also resulted in much longer training time on this task.
Scheduling $\beta$ to increase throughout training might allow for faster training.
Unless otherwise specified, we use 1000/1600 units for LSTM/SRNN, and SGD with learning rate=.002, momentum=.99, and gradient clipping=1.
We train for a maximum of 1000 epochs and use sequences of length 50 taken without overlap.
When we encounter a NaN in the cost function, we divide the learning rate by 2, and restart with the previous epoch's parameters.

For LSTMs, we either apply the norm-stabilizer penalty only to the memory cells, or only to the hidden state (in which case we remove the output tanh, as in \citep{Gers2000}).
Although \citet{Greff15} found the output tanh to be essential for good performance, removing it gave us a slight improvement in this task.
We compare to tanh and ReLU (with and without bias), with a grid search across cost weight, gradient clipping, and learning rate.
For simple RNNs, we found that the zero-bias ReLU (i.e.~ TRec \citep{ZBA} with threshold 0) gave the best performance.
The best performance for ReLU activation functions is obtained with the penalty applied.
For tanh-RNNs, the best performance is obtained without any regularization.
Results are better with the penalty than without for 9 out of 12 experiment settings.

\begin{figure}[h]
\begin{center}
\includegraphics[width=.99\columnwidth]{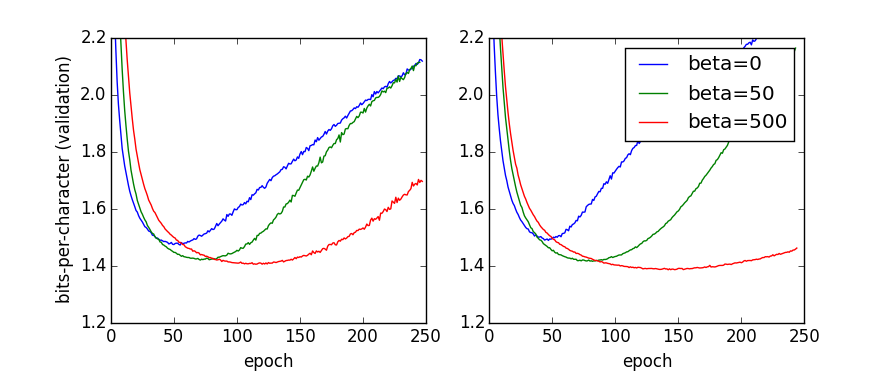}
\end{center}
\caption{Learning Curves for LSTM with different values of $\beta$.  
         Penalty is applied to the hidden state (Left), or the memory cells (Right).
         \label{fig:LSTM_lcs}}
\end{figure}

\begin{table}
  \caption{LSTM Performance (bits-per-character) on PennTreebank for different values of $\beta$.}

  \label{tab:tab_perf_LSTM}
  \begin{center}
  \begin{tabular}{ | l | c | c | c |}
    \hline
                  & $\beta=0$   & $\beta=50$   & $\beta=500$ \\ \hline
    penalize hidden state & $1.47$      & $1.41$       & $\mathbf{1.39}$    \\ \hline
    penalize memory cell & $1.49$      & $1.42$       & $\mathbf{1.40}$    \\ \hline
  \end{tabular}
  \end{center}
\end{table}

\subsubsection{Alternative costs} \label{altcost}
We compare 8 alternatives to the norm-stabilizer cost on PennTreeBank for IRNNs without biases (see Table~\ref{tab:altcost}), using the same setup as in \ref{ptb}.
These include relative error, $L_1$ norm, absolute difference, and penalties that don't target successive time-steps.
The following two penalties performed very poorly and were not included in the table: $| \Delta \norm{h_t}_2 |$, $\norm{h_t}_2^2$.

We find that our proposal of penalizing successive states' norms gives the best performance, but some alternatives seem promising and deserve further investigation.
In particular, the relative error could be more appropriate; unlike the norm-stabilizer cost, it cannot be reduced simply by dividing all of the hidden states by a constant.
The value 5 was chosen as a target for the norms based on the value found by our proposed cost; in practice it would be another hyperparameter to tune.
The success of the other regularizers which encourage ($L_2$) norm stability indicates that our inductive bias in favor of stable norms is useful.
% not included: abs diff, shrink norm %4.57%4.96%5.18%7.19

\begin{table} [h]
  \caption{Performance with and without norm-stabilizer penalty for different activation functions.
           Gradients are clipped at $1$ in the first and third, and $10^6$ in the second and fourth columns.}
  \label{tab:tab_perf_SRNN}
  \begin{center}
  \begin{tabular}{ | l | c | c | c | c |}
    \hline
                     & $lr=.002,gc=1$ & $lr=.002$ & $lr=.0002,gc=1$ & $lr=.0002$  \\ \hline
    tanh, $\beta=0$  & $1.71$ & $\mathbf{1.55}$ & $2.15$ & $2.15$\\ \hline
    tanh, $\beta=500$ & $1.57$ & $2.70$ & $1.79$ & $1.80$ \\ \hline
    ReLU, $\beta=0$  & $1.78$ & $1.69$ & $1.93$ & $1.93$ \\ \hline
    ReLU, $\beta=500$ & $1.74$ & $1.73$ & $\mathbf{1.65}$ & $2.04$ \\ \hline
    TRec, $\beta=0$  & $1.62$ & $1.63$ & $1.95$ & $1.88$ \\ \hline
    TRec, $\beta=500$ & $\mathbf{1.48}$ & $1.49$ & $1.56$ & $1.56$ \\ \hline
  \end{tabular}
  \end{center}
\end{table}

\begin{table}[h]
  \caption{Performance (bits-per-character) of zero-bias IRNN with various penalty terms designed to encourage norm stability.}
  \label{tab:altcost}
  \begin{center}
  \begin{tabular}{| l | c |  c | c | c | c | c | c |}
    \hline 
    &
    $(\Delta h_t)^2$ &
    $(\Delta \norm{h_t}_2)^2$ & 
    $(\frac{\Delta \norm{h_t}_2}{\norm{h_t}_2})^2$ & 
    $(\Delta \norm{h_t}_1)^2$ & 
    $(\norm{h}_2 - 5)^2$ &  
    $(\norm{h_0}_2 - \norm{h_T}_2)^2$
    \\ \hline
    $\beta=50$&
    $1.84$&
    &
    $1.60$&
    $2.96$&
    $1.49$&
    $3.81$
    \\ \hline
    $\beta=500$&
    $2.19$&
    $1.48$&
    $1.50$&
    $3.18$&
    $1.50$&
    $1.54$
    \\ \hline
  \end{tabular}
  \end{center}
\end{table}

\subsection{Phoneme Recognition on TIMIT} \label{timit}
We show that the norm-stabilizer improves phoneme recognition on the TIMIT dataset, outperforming networks regularized with weight noise and/or dropout.
For these experiments, we use a similar setup to the previous state of the art for an RNN on this task \citep{Graves13}, with CTC \citep{CTC} and bidirectional LSTMs with 3 layers of 500 hidden units (for each direction).  
We train with Adam \citep{Adam} using learning rate=.001 and gradient clipping=200.  
Unlike \citet{Graves13}, we do not use beam search or an RNN transducer.
We early stop after 25 epochs without improvement on the development set.

\begin{figure}[h]
\centering
\includegraphics[width=.85\columnwidth]{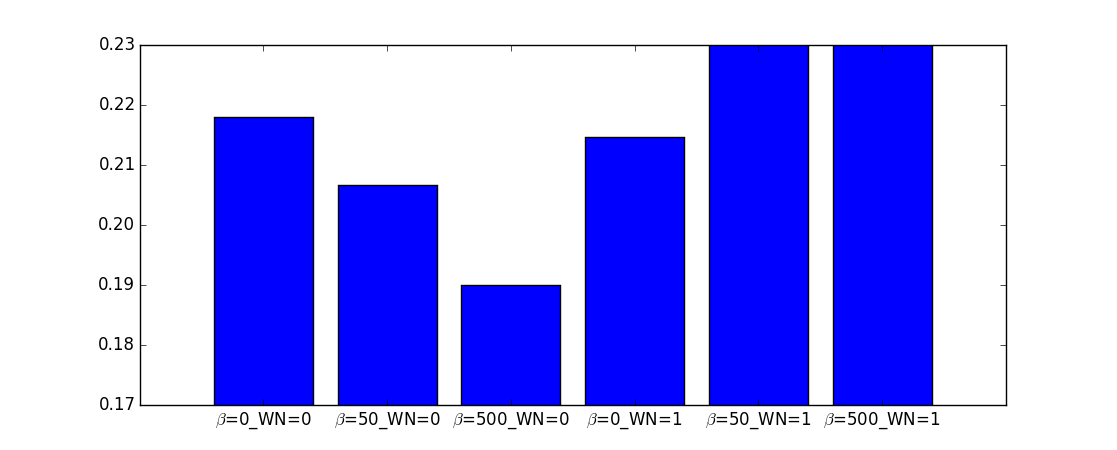}
\caption{Average PER on TIMIT core test set for different combinations or regularizers.  
         The norm-stabilizer ($\beta$) shows a clear positive effect on performance.
         Weight noise (WN) also improves performance but less so.
         Combining weight noise with norm-stabilization gives poor results.
         \label{fig:timit_bars}}
\end{figure}

We apply norm-stabilization to the hidden activations (in this case we \emph{do} use the output tanh as is standard) with $\beta \in \{0, 50,500\}$, and use standard deviation .05 for weight noise and p=.5 for dropout.  
We try all pair-wise combinations of the regularization techniques.
We run 5 experiments for each of these 10 settings, and report the average phoneme error rate (PER).
Combining weight noise and norm-stabilization gave poor performance, with some networks failing to train, these results are omitted.
Adding dropout had a minor effect on results. 
Norm-stabilized networks had the best performance (see figure~\ref{fig:timit_bars} and table~\ref{tab:TIMIT}).
% TODO: below
Inspired by these results, we decided to train larger networks with more regularization, and observed further performance improvements (see table~\ref{tab:TIMIT1000}).
We also used a higher ``patience'' for our early stopping criterion here, terminating after 100 epochs without improvement.
Unlike previous experiments, we only ran one experiment with each of these settings.
The network with 750 hidden units and $\beta=1000$ gave the best performance on the development set, with dev/test PER of 16.2\%/18.6\%.
This is competitive with the state of the art results on this task from \citet{Graves13} and we evaluate without beam search or RNN transducer. although \citet{CNN_TIMIT} achieved 13.9\%/16.7\% using convolutional neural networks.
The network with 1000 hidden units and $\beta=1000$ achieved dev/test PER of 16.7\%/17.5\%.  

%\FloatBarrier

%\vspace{-1.5cm}

\begin{table}
    \caption{Phoneme Error Rate (PER) on TIMIT for different experiment settings, average of 5 experiments.  
    Norm-stabilized networks achieve the best performance.
    The regularization parameters are: $\beta$ - norm stabilizer, $p$ - dropout probability, $\sigma$ - standard deviation of additive Gaussian weight noise.}
  \label{tab:TIMIT}
\centering
      %\begin{tabular}{ | l | p{1.5cm} | p{1.5cm} | p{1.5cm} | p{1.5cm} | p{1.5cm} | p{1.5cm} |}
      \begin{tabular}{ | l | c | c | c | c | c | c | c | c |}
    \hline
          %& $\beta=0, WN=0$ &  $\beta=50, WN=0$ & $\beta=500, WN=0$ & $\beta=0, WN=1$ &  $\beta=50, WN=1$ & $\beta=500, WN=2$ \\ \hline
    & \parbox{1cm}{$\beta=0 \\ \sigma=0 \\ p=0$} & \parbox{1.1cm}{$\beta=50 \\ \sigma=0 \\ p=0$} & \parbox{1.3cm}{$\beta=500 \\ \sigma=0 \\ p=0$} & \parbox{1.2cm}{$\beta=0 \\ \sigma=.05 \\ p=0$}  & \parbox{1.1cm}{$\beta=0 \\ \sigma=0 \\ p=.5$} & \parbox{1.1cm}{$\beta=50 \\ \sigma=0 \\ p=.5$} & \parbox{1.3cm}{$\beta=500 \\ \sigma=0 \\ p=.5$} & \parbox{1.2cm}{$\beta=0 \\ \sigma=.05 \\ p=.5$}  \\ \hline
    test  & $21.8$  & $20.7$  & $\mathbf{19.0}$  & $21.5$  & $21.9$  & $20.9$  & $19.4$  & $21.1$  \\ \hline
    dev   & $19.6$  & $18.6$  & $\mathbf{16.9}$  & $19.1$  & $19.5$  & $18.5$  & $17.0$  & $18.9$  \\ \hline
  \end{tabular}
\end{table}  

\begin{table}
  \caption{Phoneme Error Rate (PER) on TIMIT for experiments with $n$ hidden units and more norm-stabilizer regularization ($\beta$).
  Networks regularized with weight noise $\sigma=.05$ when $\beta =0$.
  The network with 750 units and $\beta=1000$ achieved the best dev PER (16.17).}
  \label{tab:TIMIT1000}
\centering
      %\begin{tabular}{ | l | p{1.5cm} | p{1.5cm} | p{1.5cm} | p{1.5cm} | p{1.5cm} | p{1.5cm} |}
      \begin{tabular}{ | l |  c | c | c | c |c | c | c | c |}
    \hline
          %& $\beta=0, WN=0$ &  $\beta=50, WN=0$ & $\beta=500, WN=0$ & $\beta=0, WN=1$ &  $\beta=50, WN=1$ & $\beta=500, WN=2$ \\ \hline
    & \parbox{1.25cm}{$\beta=0 \\ n=750$} & \parbox{1.25cm}{$\beta=500 \\ n=750$} & \parbox{1.4cm}{$\beta=1000 \\ n=750$} & \parbox{1.4cm}{$\beta=1500 \\ n=750$}  & \parbox{1.25cm}{$\beta=0 \\ n=999$} & \parbox{1.25cm}{$\beta=500 \\ n=999$} & \parbox{1.4cm}{$\beta=1000 \\ n=999$} & \parbox{1.4cm}{$\beta=1500 \\ n=999$}  \\ \hline
    test  & $21.9$  & $18.8$  & $18.6$  & $18.0$  & $21.8$  & $19.5$  & $\mathbf{17.5}$  & $18.6$  \\ \hline
    dev   & $19.6$  & $16.8$  & $\mathbf{16.2}^*$  & $\mathbf{16.2}$  & $19.1$  & $17.4$  & $16.7$  & $16.7$  \\ \hline
  \end{tabular}
\end{table}  

%\vspace{-1cm}

%\FloatBarrier

\subsection{Adding Task} \label{adding}
The adding task \citep{LSTM} is a toy problem used to test an RNN's ability to model long-term dependencies.
The goal is to output the sum of two numbers seen at random time-steps during training; inputs at other time-steps carry no information.
Each element of an input sequence consists of a pair $\{n, i\}$, where $n \in [0,1]$ is chosen at uniform random and $i \in \{0,1\}$ indicates which two numbers to add.
We use sequences of length 400.
In \citet{IRNN}, none of the models were able to reduce the cost below the ``short-sighted'' baseline set by predicting the first (or second) of the indicated numbers (which gives an expected cost of $\frac{1}{12}$) for this sequence length.
We are able to solve this task more successfully.
We use uniform initialization in $[-.01, .01]$, learning rate=.01, gradient clipping=1.
We compare across nine random seeds with and without the norm-stabilizer (using $\beta = 1$).
The norm-stabilized networks reduced the test cost below $\frac{1}{12}$ in 8/9 cases, averaging .059 MSE.
The unregularized networks averaged .105 MSE, and only outperformed the short-sighted baseline in 4/9 cases, also failing to improve over a constant predictor in 4/9 cases.

\subsection{Visualizing the effects of norm-stabilization}
To test our hypothesis that stability helps networks generalize to longer sequences than they were trained on, we examined the costs and hidden norms at each time-step.

\begin{figure}[h]
\centering
\includegraphics[width=.95\columnwidth]{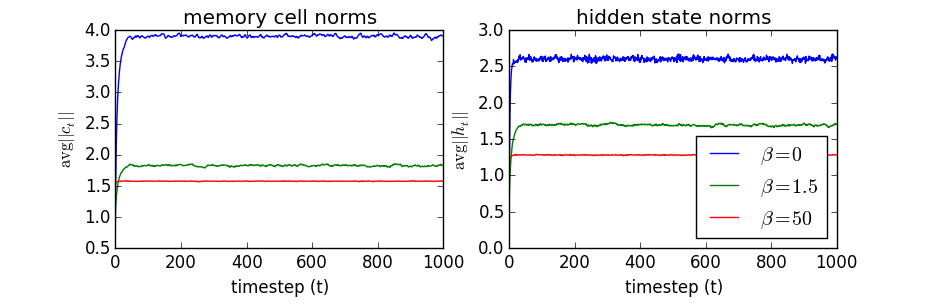}
\caption{Norm (y-axis) of LSTM memory cells (Left) and hidden states (Right) for different values of $\beta$, across time-steps (x-axis).
         Non-zero values dramatically reduce the mean and variance of the norms.
         LSTM memory cells have the potential to grow linearly, but instead exhibit natural stability.
         \label{fig:LSTM_activations}}
\end{figure}

\begin{figure}[h]
%\begin{center}
\centering
\includegraphics[width=.80\columnwidth]{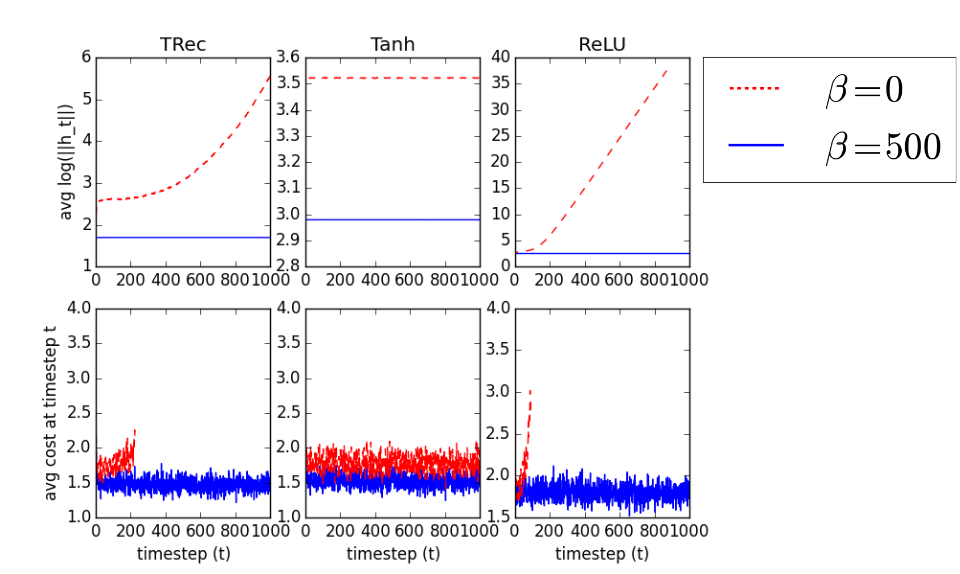}
%\end{center}
\caption{Top: average logarithm of hidden norms as a function of time-step. 
         Bottom: average cost as a function of time-step.
         Solid blue - $\beta=500$, dashed red - $\beta=0$.  
         Notice that IRNN's activations explode exponentially (linearly in the log-scale) within the training horizon, causing cost quickly go to infinity outside of the training horizon (50 time-steps).
         \label{fig:exploding}}
\end{figure}

Comparing identical SRNNs trained with and without norm-stabilizer penalty, we found LSTMs and RNNs with tanh activation functions continued to perform well far beyond the training horizon.
Although the activations of LSTM's memory cells could potentially grow linearly, in our experiments they are stable.  
Applying the norm-stabilizer does significantly decrease their average norm and the variability of the norm, however (see figure~\ref{fig:LSTM_activations}).  
IRNNs, on the other hand, suffered from exploding activations, resulting in poor performance, but the norm-stabilizer effectively controls the norms and maintains a high level of performance; see figure~\ref{fig:exploding}.
Norm-stabilized IRNNs' performance and norms were both stable for the longest horizon we evaluated (10,000 time-steps).

%\begin{figure}[h]
%\begin{minipage}{.45\textwidth}
%\includegraphics[height=4.5cm]{lstm_hids.png}
%\end{minipage} 
%\hfill
%\begin{minipage}{.45\textwidth}
%\includegraphics[height=4.5cm]{lstm_mems.png}
%\end{minipage}
%\caption{Norm (y-axis) of LSTM hidden states (Left) and memory cells (Right) for different values of $\beta$, across time-steps (x-axis).
%         The blue curve at the top is when $\beta = 0$.  Non-zero values dramatically reduce the mean and variance of the norms.
%         LSTM memory cells have the potential to grow linearly, but instead exhibit natural stability.
%         \label{fig:LSTM_activations}}
%\end{figure}

\begin{figure}[h]
\includegraphics[height=4.8cm]{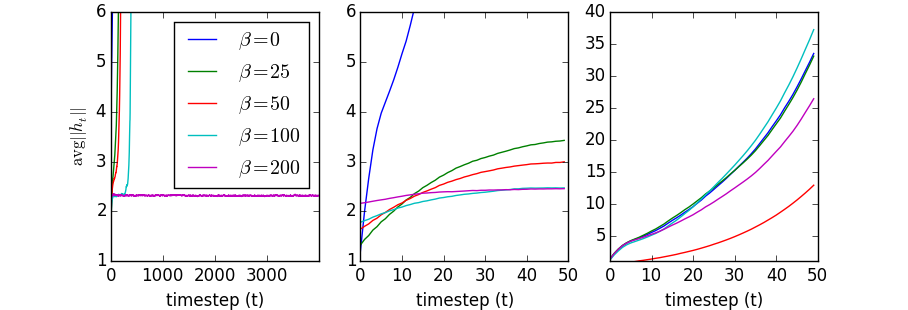}
\caption{Hidden norms as a function of time-step for values from 0 to 200 of the norm-stabilizer (Left and Center) vs. a penalty on the initial and final norms (Right).
         The norm-stabilizer delays the explosion of activations by changing the shape of the curve, extending the flat region.
         \label{fig:compare_blowups}}
\end{figure}

For more insight on why the norm-stabilizer outperforms alternative costs, we examined the hidden norms of networks trained with values of $\beta$ ranging from 0 to 200 on a dataset of 1000 length-50 sequences taken from wikipedia \citep{enwiki}.
When we penalize the difference of the initial and final norms, or the difference of the norms from some fixed value, increasing the cost does not change the shape of the norms; they still begin to explode within the training horizon (see figure~\ref{fig:compare_blowups}).
For the norm-stabilizer, however, increasing the penalty significantly delayed (but did not completely eradicate) activation explosions on this dataset.

\begin{figure}[h]
\begin{minipage}{.6\columnwidth}
\includegraphics[height=3cm]{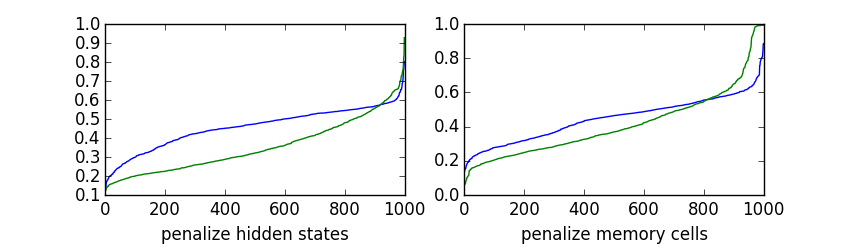}
\end{minipage} \hfill
\begin{minipage}{.2\columnwidth}
\includegraphics[height=3cm]{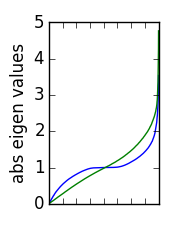}
\end{minipage}
\caption{Left: sorted distribution of average forget-gates for different memory cells in LSTM.
         Right: sorted absolute value of eigenvalues of $W_{hh}$ in IRNN.
         Blue - $\beta=0$, Green - $\beta=500$
         \label{fig:eigs}}
\end{figure}

We also noticed that the distribution of activations was more concentrated in fewer hidden units when applying norm-stabilization on PennTreebank.
Similarly, we found that the forget gates in LSTM networks had a more peaked distribution (see figure~\ref{fig:eigs}), while the average across dimensions was lower (so the network was forgetting more on average at each time step, but a small number of units were forgetting less).
Finally, we found that the eigenvalues of regularized IRNN's hidden transition matrices had a larger number of large eigenvalues, while the \emph{unregularized} IRNN had a much larger number of eigenvalues closer to $1$ in absolute value (see figure~\ref{fig:eigs}).
This supports our hypothesis that orthogonal transitions are not inherently desirable in an RNN.
By explicitly encouraging stability, the norm-stabilizer seems to favor solutions that maintain stability via selection of active units, rather than restricting the choice of transition matrix.

\FloatBarrier

%\begin{figure}[h]
%\begin{minipage}{.8\columnwidth}
%\begin{center}
%\includegraphics[height=.6cm]{legend_compare_blowups.png}
%\end{center}
%\end{minipage} 
%\vfill
%\begin{minipage}{.8\columnwidth}
%\includegraphics[height=3.3cm]{compare_blowups.png}
%\end{minipage}
%\caption{Hidden norms as a function of time-step for values from 0 to 400 of the norm-stabilizer (Left and Center) vs. a penalty on the initial and final norms (Right).
%         The norm-stabilizer delays the explosion of activations by changing the shape of the curve, extending the flat region.
%         \label{fig:compare_blowups}}
%\end{figure}

%%%%%%%%%%%%%%%%%%%%%%%%%%%%%%%%%%%%%%%%%%%%%%%%%%%%%%%%%%
\FloatBarrier
\section{Conclusion}
We introduced norm-based regularization of RNNs to prevent exploding or vanishing \emph{activations}. 
We compare a range of novel methods for encouraging or enforcing norm stability.  
The best performance is achieved by penalizing the squared difference of subsequent hidden states' norms.  
This penalty, the \emph{norm-stabilizer}, improved performance on the tasks of language modeling and addition tasks, and gave state of the art RNN performance on phoneme recognition on the TIMIT dataset.

Future work could involve:
\begin{itemize}
\item Exploring the relationship between stability and generative modeling with RNNs
\item Applying norm-regularized IRNNs to more challenging tasks
\item Applying similar regularization techniques to feedforward nets
\end{itemize}

\subsubsection*{Acknowledgments}
This research was developed with funding from the Defense Advanced Research Projects Agency (DARPA) and the Air Force Research Laborotory (AFRL) . The views, opinions and/or findings expressed are those of the authors and should not be interpreted as representing the official views or policies of the Department of Defense or the U.S. Government.
We appreciate the many k80 GPUs provided by ComputeCanada.
The authors would like to thank the developers of Theano \citep{Theano} and Blocks \citep{Blocks}.
Special thanks to Alex Lamb, 
Amar Shah, 
Asja Fischer, 
Caglar Gulcehre, 
Cesar Laurent, 
Dmitriy Serdyuk, 
Dzmitry Bahdanau, 
Faruk Ahmed, 
Harm de Vries, 
Jose Sotelo, 
Marcin Moczulski, 
Martin Arjovsky, 
Mohammad Pezeshki, 
Philemon Brakel, 
and Saizhen Zhang for useful discussions and/or sharing code.

\bibliography{mybib}
\bibliographystyle{iclr2016_conference}

\end{document}